\documentclass[10pt,a4paper]{article}
\usepackage[top=0.85in,left=1.5in,right=2.0in,footskip=0.75in]{geometry}

\usepackage[utf8]{inputenc}

\usepackage{cite}

\usepackage{nameref,hyperref}

\usepackage[right]{lineno}

\usepackage{microtype}
\DisableLigatures[f]{encoding = *, family = * }

\usepackage{changepage}

\usepackage[aboveskip=1pt,labelfont=bf,labelsep=period,singlelinecheck=off]{caption}

\makeatletter
\renewcommand{\@biblabel}[1]{\quad#1.}
\makeatother

\usepackage{lastpage,fancyhdr,graphicx}
\usepackage{epstopdf}
\fancyheadoffset[]{0.5in}
\fancyfootoffset[]{0.5in}

\usepackage{color}

\definecolor{Gray}{gray}{.25}

\usepackage{graphicx}

\usepackage{sidecap}

\usepackage{wrapfig}
\usepackage[pscoord]{eso-pic}
\usepackage[fulladjust]{marginnote}
\reversemarginpar

\usepackage{amssymb}
\usepackage{amsmath}
\usepackage{mathtools}
\usepackage{mathrsfs}
\usepackage{algorithm}
\usepackage{varwidth}
\usepackage[noend]{algpseudocode}
\usepackage{bm}
\usepackage{multirow}
\usepackage{booktabs}
\usepackage{natbib}
\usepackage{float}

\newcommand{\norm}[1]{\left\lVert#1\right\rVert}
\newcommand{\bs}{\boldsymbol}
\newcommand{\argmax}{\arg\max}

\begin{document}
\vspace*{0.35in}

\begin{flushleft}
	{\Large
		\textbf\newline{Boredom-driven curious learning by Homeo-Heterostatic Value Gradients.}
	}
	\newline
	\\
	Yen Yu\textsuperscript{1*},
	Acer Y.C. Chang\textsuperscript{1},
	Ryota Kanai\textsuperscript{1},
	\\
	\bigskip
	\bf{1} Araya, Inc., Tokyo, Japan
	\\
	\bigskip
	* Corresponding author: Yen Yu (first.lastname@araya.org)
\end{flushleft}

\section*{Abstract}

This paper presents the Homeo-Heterostatic Value Gradients (HHVG) algorithm as a formal account on the constructive interplay between boredom and curiosity which gives rise to effective exploration and superior forward model learning. We envisaged actions as instrumental in agent's own epistemic disclosure. This motivated two central algorithmic ingredients: devaluation and devaluation progress, both underpin agent's cognition concerning intrinsically generated rewards. The two serve as an instantiation of homeostatic and heterostatic intrinsic motivation. A key insight from our algorithm is that the two seemingly opposite motivations can be reconciled---without which exploration and information-gathering cannot be effectively carried out. We supported this claim with empirical evidence, showing that boredom-enabled agents consistently outperformed other curious or explorative agent variants in model building benchmarks based on self-assisted experience accumulation. 

\section{Introduction}

In this study, we argue that action is instrumental in epistemic disclosure in and of agent itself. The implication of this statement is twofold: (1) for agents whose innate goal appeals to their own knowledge gain, the occurrence of curiosity rests upon the presence of devaluation (and hence goal-directedness); (2) boredom---consequential to devaluation---and curiosity entail a mutually reinforcing cycle for such kind of disclosure to ensue.

Animal studies have shown that learning stimulus-response (S-R) association through action-outcome reinforcement is but one facet of instrumental behaviour. Internally, animals may build models that assign values to reappraise experienced outcomes. This expands the landscape of instrumental behaviour to include stimulus-outcome-response (S-O-R) learning system---or goal-directed learning \citep{Balleine.1998}. Goal-directed behaviour is known in both empirical and computational approaches to support adaptive and optimal action selection \citep{adams1981instrumental, adams1982variations, mannella2016goal}. Central to such behavioural adaptiveness is devaluation. This means for a given action-outcome pair the associated reinforcing signal is no longer monotonic. Instead, outcome value varies under reappraisal within according to their relevance to or attainment of goal. 

One classic paradigm of devaluation that shapes agent's behavioural pattern is of food via satiation. In the context of epistemic disclosure, an analogy can be drawn between devaluation and the emergence of boredom, in which one's assimilation of knowledge reduces the value of similar knowledge in future encounter. The relationship between boredom and outcome devaluation has a long history in psychological research. Empirical findings indicated that boredom is reportedly accompanied by negative affective experience, suggesting that experienced outcomes are intrinsically evaluated and considered as less valuable \cite{bench2013function, van2012boredom, fahlman2009does, perkins1985cognitive, vodanovich1991boredom}. Psychophysiological studies also demonstrated that boredom plays an active role of information-seeking behaviour. Subjects showing higher levels of reported boredom are accompanied by increased autonomic arousal, such as heart rate and galvanic skin response. These findings is in line with our key notion that boredom intrinsically and actively drives agents learning behaviours \citep{london1972increase, harris2000correlates}. Consistent with our argument, evidence also showed that boredom is associated with increase in creativity \cite{harris2000correlates, schubert1977boredom, schubert1978creativity}. This suggests that the presence of boredom serves to reconfigure agent's instrumental device in order to escape devalued states.

Curiosity, irrespective of being a by-product of external goal-attainment or an implicit goal in and of agent itself, is often ascribed to as a correlate of information-seeking behaviour \citep{2013gottlieb}. Behaviours exhibiting curious quality are observed in humans and animals alike, suggesting an universal role of curiosity in shaping one's fitness. Though the exact neural mechanism underlying the emergence of curious behaviour still remains obscure, current paradigms have their focus on (1) novelty disclosure and (2) uncertainty reduction aspects of information-seeking \citep{pathak2017icm, friston2017curiosity, bellemare2016count, ostrovski2017count}. Indeed, both aspects can be argued to improve agent's fitness in epistemic landscape if the agent elects to incorporate the novelty or uncertainty. 

In intrinsic motivation literature \citep{intrinsicmotiv}, although one can readily associate boredom with homeostatic motivation and curiosity with heterostatic motivation, our argument suggests they can in fact be complementary. Our contribution thus pertains to the reconciliation of homeo-heterstatic motivations.

\section{Markov Decision Process}

In what follows, we briefly review preliminaries for the ensuing algorithm. We focus on well-established themes surrounding typical reinforcement learning, including Markov Decision Process and value gradients as a policy optimisation technique. 

In Markov Decision Process (MDP) one considers the tuple $(S, A, R, P, \pi, \gamma)$. $S$ and $A$ are spaces of real vectors whose member, $\bs{s} \in S$ and $\bs{a} \in A$, represent states (or sensor values) and actions. $R$ is some reward function defining the mapping $R: S \times A \to \mathbb{R}$. The probabilities associated with states and actions are given by the forward model $P (S^\prime|S=\bs{a}, A=\bs{s})$ and the action policy $\pi(A|S=\bs{s})$. Throughout the paper we use the `prime' notation, e.g., $\bs{s}^\prime$, to represent one time step into the future: $\bs{s}^\prime=\bs{s}(t+1)$.

The goal of MDP is to optimally determine the action policy $\pi^\ast$ such that the expected cumulative reward over a finite (or infinite) horizon is maximised. Considering a finite horizon problem with discrete time, $t \in [0, T]$, this is equivalent to $\pi^\ast = \argmax_\pi \mathbb E_{\bs{a} \sim \pi}\left[ \sum_{t=0}^{T}\gamma^t R(\bs{s}(t), \bs{a}(t)) \right]$, where $\gamma \in [0, 1]$ is the discount factor.

Many practical approaches for solving MDP often resort to approximating state-action value $q(\bs{a}, \bs{s})$ or state value $v(\bs{s})$ functions \citep{sutton1998reinforcement, deepmind2013atari, deepmind2015lillicrap, svg}. These value functions are given in the Bellman equation
	\begin{equation}
	\begin{aligned} \label{eq:bellman}
	v(\bs{s}) 
	&=
	\mathbb{E}_{\pi(\bs{a}|\bs{s})} 
	\Bigl[ R(\bs{a}, \bs{s}) + \gamma q(\bs{a}, \bs{s}) \Bigr] \\
	&=
	\mathbb{E}_{\pi(\bs{a}|\bs{s})} 
	\Bigl[ R(\bs{a}, \bs{s}) + \gamma \mathbb{E}_{P(\bs{s}^\prime|\bs{a}, \bs{s})} [v(\bs{s}^\prime)] \Bigr]
	\end{aligned}
	\end{equation}
When differentiable forward model and reward function are both available, policy gradients can be analytically estimated using value gradients \citep{2012fairbank, svg}.

\section{Homeo-Heterostatic Value Gradients} \label{sec:algorithm}

This section describes formally the algorithmic structure and components of the Homeo-Heterostatic Value Gradients, or HHVG. The naming of HHVG suggests its connections with homeostatic and heterostatic intrinsic motivations \citep{intrinsicmotiv}. A homeostatic motivation encourages an organism to occupy a set of predictable, unsurprising states. Whereas, a heterostatic motivation does the opposite; curiosity belongs to this category. 

The algorithm offers reconciliation between the two seemingly opposite qualities and concludes with their cooperative nature. Specifically, the knowledge an organism maintains about its homeostatic boundary helps instigate outbound heterostatic drives. In return, satisfying heterostatic drives broadens the organism's boundary of comfort. As a consequence, the organism not only improves its fitness in terms of homeostatic outreach but also becomes effectively curious.

It is instructive to overview the nomenclature of the algorithm. We consistently associate homeostatic motivation with the emergence of {\it boredom}, which reflects the result of having incorporated novel information into one's knowledge, thereby diminishing the novelty to begin with. This is conceptually compatible with outcome {\it devaluation} or induced satiety in instrumental learning. {\it Devaluation progress} is therefore referred to as one's epistemic achievement. That is, the transitioning of a priori knowledge to one of having assimilated otherwise unknown information. The devaluation progress is interpreted as an instantiation of intrinsic reward. The drive to maintain steady rewards conforms to a heterostatic motivation.

\begin{figure}[H]
	\begin{center}
		\includegraphics[width=9cm]{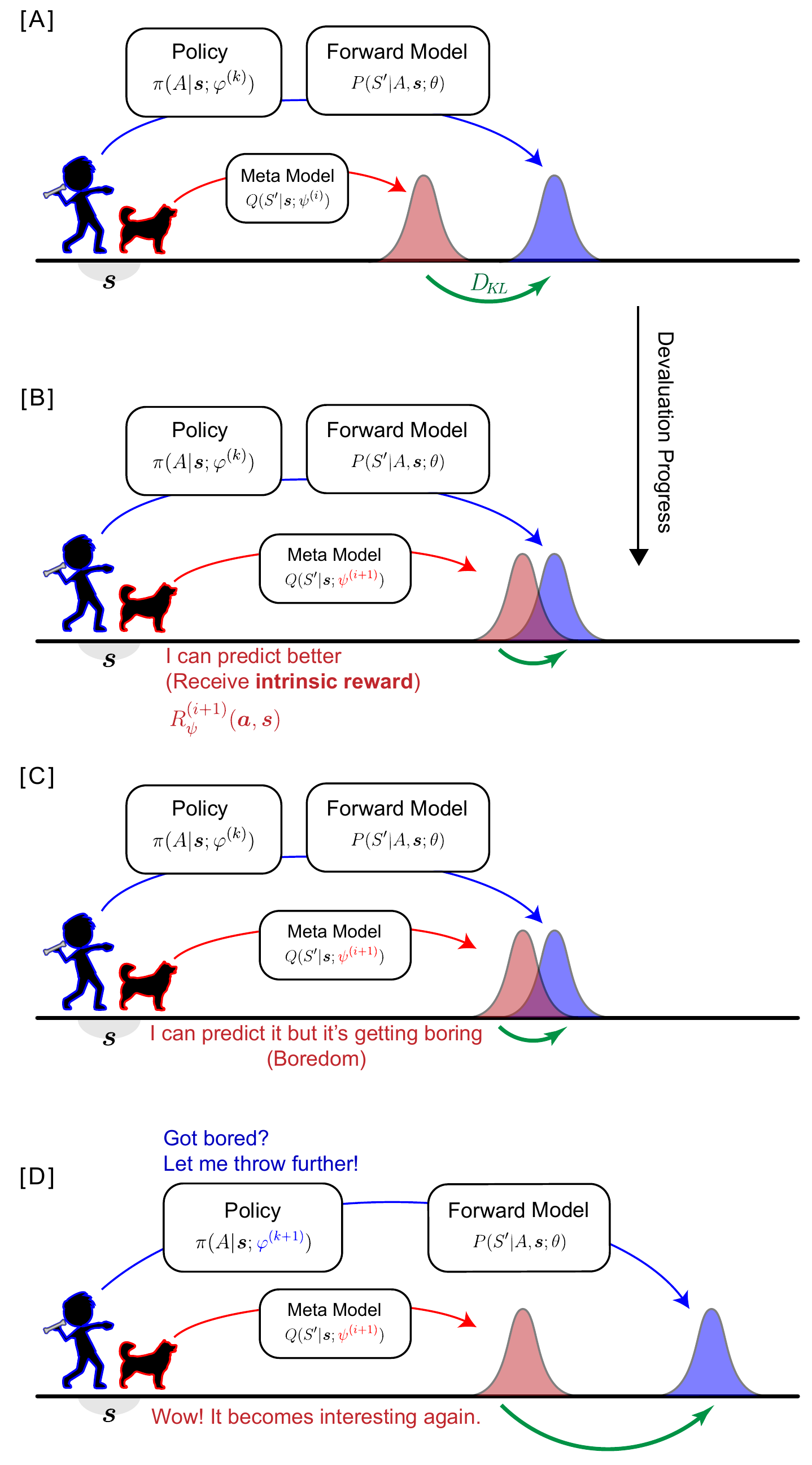}
	\end{center}
	\caption{Intuitive understanding of the Homeo-Heterostatic Value Gradients (HHVG) algorithm. [A] The algorithm can be interpreted as the cooperative interplay between a thrower (kid; blue) and a catcher (dog; red). The thrower is equipped with a forward model that estimates its aiming and is controlled by an action policy. Without knowing the thrower's policy, the catcher (meta-model), in order to make good catches, infers where the thrower is aiming on average. [B] The catcher is interested in novel, unpredicted throws. Whenever the catcher improves its predictive power some intrinsic reward (devaluation progress) is generated. [C] As the catcher progresses further, similar throws become highly predictable, thus inducing a sense of boredom. [D] To make the interplay interesting again, the thrower is driven to devise new throws, so that the catcher can afford to make further progress. By repeating [A--B] the thrower has attempted diverse throws and known well about its aim. At the same time, the catcher will assume a vantage point for any throw.}
	\label{fig:intuition}
\end{figure}

An intuitive understanding of HHVG is visualised in Figure \ref{fig:intuition}. Imagine the interplay between a thrower and their counterpart --- a catcher. The catcher anticipates where the thrower is aiming and makes progress by improving its prediction. The thrower, on the other hand, keeps the catcher engaged by devising novel aims. Over time, the catcher knows well what the thrower is capable of, whilst the thrower has attempted a wide spectrum of pitches.

In the algorithm, the thrower is represented by a forward model attached to a controller (policy) and the catcher a ``meta-model''. We unpack and report them individually. Procedural information is summarised in Algorithm \ref{alg:hhvg}.

\subsection{Forward model} \label{subsec:fm}

We start by specifying at current time the state and action sample as $\bs{s}$ and $\bs{a}$. The forward model describes the probability distribution over future state $S^\prime$, given $\bs{s}$, $\bs{a}$, and parameter $\theta$.
	\begin{equation}
	\begin{aligned} \label{eq:fm}
	P(S^\prime | A=\bs{a}, S=\bs{s}; \theta)
	\end{aligned}
	\end{equation}
The entropy associated with $S^\prime$, conditioned on $\bs{s}$ and $\bs{a}$, gives a measure of the degree to which $S^\prime$ is informative on average. We referred to this measure as one of {\it interestingness}. Note this is a different concept from the `interestingness' proposed by \citet{schmidhuber2008driven}, which is the first-order derivative of compressibility.

\subsection{Boredom, outcome devaluation, and meta-model} \label{subsec:boredom}

Boredom, in common understanding, is perhaps not unfamiliar to most under the situation of being exposed to certain information which one has known well by heart. It is the opposite of being interested. In the current work, we limited the exposure of information to those being disclosed by one's actions.

To mark the necessity of boredom, we first identify the limitation of a naive instantiation of curiosity; then, we show that the introduction of boredom serves to resolve this limitation.

Consider the joint occurrence of future state $S^\prime$ and action $A$: $ P (S^\prime, A|S=\bs{s}; \theta, \varphi)$. This is derived from product rule given Equation \ref{eq:fm} and action policy $\pi(A|S=\bs{s}; \varphi)$, parametrised by $\varphi$ (action policy is revisited in Section \ref{subsec:policy}). 

A naive approach to curiosity is by optimising the action policy, such that $A$ is predictive of maximum {\it interestingness} (see Section \ref{subsec:fm}) about the future.

However, this approach would certainly lead to the agent behaving habitually and, as a consequence, becoming obsessive about a limited set of outcomes. In other words, a purely interestingness-seeking agent is a darkroom agent (see Section\ref{subsec:homeohetero}; also \citet{darkroom} for related concept).

The problem with the naively curious agent is that it perceives novelty as permanently novel. The agent has no recourse to inform itself via assimilating the information that brought about novelty. If the agent is otherwise endowed with the assimilation capacity, a sense of boredom would be induced. The induction of boredom essentially causes the agent to value the same piece of information less, thus changing the agent's perception towards interestingness. If the agent were to pursue the same interestingness-seeking policy, a downstream effect of boredom would drive the agent to seek out other information that could have been known. This conception amounts to an implicit goal of {\it devaluating} known outcomes.

To this end, we introduce the following meta-model $Q$ to represent {\it a priori} knowledge about the future. The meta-model, parametrised by $\psi$, is an approximation to the {\it true} marginalisation of joint probability $P(S^\prime, A|S=\bs{s}; \theta, \varphi)$ over $A$:
	\begin{equation}
	\begin{aligned} \label{eq:mm}
	Q(S^\prime | S=\bs{s}; \psi) & \approx 
		P (S^\prime | S=\bs{s}; \theta, \varphi) \\
		&= \sum_A \Big[ P(S^\prime, A|\bs{s}; \theta, \varphi) \Bigr] \\
		&= \sum_A \Bigl[ P (S^\prime| A, \bs{s}; \theta) \pi(A|\bs{s}; \varphi) \Bigr] \\
	\end{aligned}
	\end{equation}

We associate the occurrence of boredom, or, synonymously, outcome devaluation, with minimising the devaluation objective with respect to $\psi$. The devaluation objective is given by the Kullback-Leibler (KL) divergence:
	\begin{equation}
	\begin{aligned} \label{eq:mm-loss}
	\mathcal{L}_{mm}(\psi) &:=
	\mathcal{L}(\bs{a}, \bs{s}; \psi, \theta) \\
		&\phantom{:}= 
		D_{KL}\left[ P(\bs{s}^\prime|\bs{a}, \bs{s}; \theta) 
		\middle\Vert 
		Q(\bs{s}^\prime|\bs{s}; \psi) \right]
	\end{aligned}
	\end{equation}
For the notation $\mathcal{L}_{mm}(\psi)$, where $mm$ stands for meta-model, we have dropped the dependence of $\theta$, $\bs{s}$, and $\bs{a}$. This only serves to emphasise that optimising the devaluation objective is with respect to $\psi$.

\subsection{Devaluation progress, intrinsic reward, and value learning}

Through the use of KL-divergence in Equation \ref{eq:mm-loss}, we emphasise the complementary nature of devaluation in relation to a knowledge-gaining process. That is to say, devaluation results in information gain for the agent. This, in fact, can be regarded as cognitively rewarding and, thus, serves to motivate our definition of intrinsic reward. 

One rewarding scenario happens when $Q(S^\prime|\bs{s}; \psi)$ has all the information there is to be possessed by $A$ about $S^\prime$. $A$ is therefore rendered redundant. One may speculate, at this point, the agent could opt for inhibiting its responses. Disengaging actions potentially saves energy which is rewarding in biological sense. 

Alternatively, the agent may attempt to develop new behavioural repertoires, brining into $S^\prime$ new information (i.e., novel outcomes) that is otherwise unknown to $Q$. The ensuing sections will focus on this line of thinking.

From Equation \ref{eq:mm-loss}, we construct the quantity {\it devaluation progress} to represent an intrinsically motivated reward. The devaluation progress is given by the difference between KL-divergences before and after devaluation (as indicated by the superscript $(i+1)$):
	\begin{equation}
	\begin{aligned} \label{eq:deval-prog}
	R_\psi^{(i+1)}(\bs{a}, \bs{s}) 
		&:= 
		\mathcal{L}(\bs{a}, \bs{s}; \psi^{(i)}, \theta) - 
		\mathcal{L}(\bs{a}, \bs{s}; \psi^{(i+1)}, \theta) \\
		&\phantom{:}=	
		\mathcal{L}_{mm}(\psi^{(i)}) - \mathcal{L}_{mm}(\psi^{(i+1)}),
	\end{aligned}
	\end{equation}
Here, we write $R_\psi^{(i+1)}(\bs{a}, \bs{s})$ in accordance with notational convention in reinforcement learning, where reward is typically a function of state and action. Subscript $\psi$ indicates the dependence of $R$ on meta model parameter.

Having established the intrinsic reward, value learning is such that the value function approximator $\hat{V}(\bs{s}; \nu)$ follows the Bellman equation $V(\bs{s}) = \mathbb{E}_{\bs{a}}[R(\bs{a}, \bs{s}) + \gamma \mathbb{E}_{\bs{s}^\prime}[V(\bs{s}^\prime)]]$. In practice, we minimise the objective with respect to $\nu$:
	\begin{equation}
	\begin{aligned} \label{eq:vf-loss}
	\mathcal{L}_{vf}(\nu) 
		&:= 
		\mathcal{L}(\bs{s}^\prime, \bs{a}, \bs{s}; \nu) \\ 
		&\phantom{:}= 
		\norm{y - \hat{V}(\bs{s}; \nu) }^2 \\
		y 
		&\phantom{:}= 
		R_\psi^{(i+1)}(\bs{a}, \bs{s}) + \gamma \hat{V}(\bs{s}^\prime; \tilde\nu)
	\end{aligned}
	\end{equation}

\subsection{Policy optimisation} \label{subsec:policy}

We define action policy at state $S=\bs{s}$ as the probability distribution over $A$ with parameter $\varphi$:
	\begin{equation}
	\begin{aligned} \label{eq:policy}
	\pi(A|S=\bs{s}; \varphi)
	\end{aligned}
	\end{equation}

Our goal is to determine the policy parameter $\varphi$ that maximises the expected sum of future discounted rewards. One approach is by applying Stochastic Value Gradients \citep{svg} and maximises the value function. We thus define our policy objective as follows (notice the negative sign; we used a gradient update rule that defaults to minimisation):
	\begin{equation}
	\begin{aligned} \label{eq:policy-loss}
	\mathcal{L}_{ap}(\varphi) 
		&:=
		\mathcal{L}(\bs{s}^\prime, \bs{a}, \bs{s}; \theta, \psi^{(i)}, \psi^{(i+1)}, \nu, \varphi) \\
		&\phantom{:}= - 
		\mathbb{E}_{\bs{a} \sim \pi(\cdot | \bs{s}; \varphi)} 
		\Bigl[ 
		R_\psi^{(i+1)}(\bs{a}, \bs{s}) + \gamma 
		\mathbb{E}_{\bs{s}^\prime \sim P(\cdot |\bs{a}, \bs{s}; \theta)}
		\bigl[ \hat{V}(\bs{s}^\prime; \nu) \bigr] 
		\Bigr] \\
	\end{aligned}
	\end{equation}

\subsection{Remarks on homeostatic and heterostatic regulations} \label{subsec:homeohetero}

\citet{intrinsicmotiv} outlined the distinctions between two important classes of intrinsic motivation: homeostatic and heterostatic. A homeostatic motivation is one that can be satiated, leading to certain equilibrium behaviourally; whereas a heterostatic motivation topples the agent, thus preventing it from occupying habitual states.

Our algorithm entails regulations relating to both classes of intrinsic motivation. Specifically, the devaluation objective (Equation \ref{eq:mm-loss}) realises the homeostatic aspect due to its connection with induced satiety. On the other hand, the devaluation progress (Equation \ref{eq:deval-prog}) introduced for policy optimisation instantiates a heterostatic drive to agent's behavioural pattern.

Heterostasis is motivated by the agent pushing itself towards novelty and away from devalued, homeostatic states (Equation \ref{eq:hetero}). We develop this statement more formally by first re-examining Equation \ref{eq:policy-loss}, with reference to Equation \ref{eq:deval-prog} and \ref{eq:mm-loss}. We arrived at the following form by admitting expected KL-divergence:
	\begin{equation}
	\begin{aligned} 
	&\phantom{={}} 
		-\mathbb{E}_{\bs{a} \sim \pi(\cdot | \bs{s};\varphi)} 
		\Big[ 
		D_{KL}[P(\bs{s}^\prime|\bs{a}, \bs{s}; \theta) \Vert Q(\bs{s}^\prime|\bs{s};\psi^{(i  )})] -
		D_{KL}[P(\bs{s}^\prime|\bs{a}, \bs{s}; \theta) \Vert Q(\bs{s}^\prime|\bs{s};\psi^{(i+1)})]
		\Big] \\
	&\phantom{={}} 
		- \mathbb{E}_{\bs{a}\sim\pi(\cdot|\bs{s};\varphi)}
		  \mathbb{E}_{\bs{s}^\prime \sim P(\cdot | \bs{a}, \bs{s}; \theta)} 
		  \Big[ V(\bs{s}^\prime; \nu) \Big] \\
	&= 
		- \Bigl\{ 
		I(S^\prime : A|S=\bs{s}; \psi^{(i)}, \varphi, \theta) -  
		I(S^\prime : A|S=\bs{s}; \psi^{(i+1)}, \varphi, \theta) 
		\Bigr. \\
	&\phantom{={}-\Big.} 
		+ \Bigl. 
		\mathbb{E}_{\bs{a}\sim\pi(\cdot|\bs{s};\varphi)}
		\mathbb{E}_{\bs{s}^\prime \sim P(\cdot | \bs{a}, \bs{s}; \theta)} 
		\big[ V(\bs{s}^\prime; \nu) \big] 
		\Bigr\}
	\end{aligned}
	\end{equation}

Notice that the expected devaluation progress becomes the difference between conditional mutual information $I$ before ($\psi^{(i)}$) and after devaluation ($\psi^{(i+1)})$.

Assume, for the moment, that the agent is equipped with devaluation capacity only. In other words, we replace the devaluation progress and fall back on devaluation objective, $R := \mathcal{L}_{mm}(\psi)$ (cf. Equation \ref{eq:deval-prog}). The agent is now interestingness-seeking with homeostatic regulation. We further suppose that the dynamics of $\psi$ and $\varphi$ evolve in tandem, which gives
	\begin{equation}
	\begin{aligned} \label{eq:pardyn}
	I(S^\prime : A | S=\bs{s}; \psi^{(i)}, \varphi^{(k)}) &\to I(S^\prime : A | S=\bs{s}; \psi^{(i+1)}, \varphi^{(k)}) \\
		&\to I(S^\prime : A | S=\bs{s}; \psi^{(i+1)}, \varphi^{(k+1)}) \\
		&\to I(S^\prime : A | S=\bs{s}; \psi^{(i+2)}, \varphi^{(k+1)}) \to \dots
	\end{aligned}
	\end{equation}
In practice, the nature of devaluation and policy optimisation often depends on replaying agent's experience. Taking turn applying gradient updates to $\psi$ and $\varphi$ creates a self-reinforcing cycle that drives the policy to converge towards a point mass. For instance, if the policy is modelled by some Gaussian distribution, this updating scheme would result in infinite precision (zero spread).

For curiosity, however, such parameter dynamics should not be catastrophic if we subsume the homeostatic regulation and ensure the preservation of the relation given in Equation \ref{eq:mutinfo-relation}:
	\begin{equation}
	\begin{aligned} \label{eq:mutinfo-relation}
	I(S^\prime : A | S=\bs{s}; \psi^{(i+1)}, \varphi^{(k)}) \le
		I(S^\prime : A | S=\bs{s}; \psi^{(i)}, \varphi^{(k)}) &\le
		I(S^\prime : A | S=\bs{s}; \psi^{(i+1)}, \varphi^{(k+1)}) \\
	\Rightarrow
		-I(S^\prime : A | S=\bs{s}; \psi^{(i+1)}, \varphi^{(k)}) + 
		 I(S^\prime : A | S=\bs{s}; \psi^{(i)}, \varphi^{(k)}) &\le
		 I(S^\prime : A | S=\bs{s}; \psi^{(i+1)}, \varphi^{(k+1)})
	\end{aligned}
	\end{equation}
This equation holds because the devaluation process on average has a tendency to make $A$ less informative about $S^\prime$, after which $A$ is perturbed to encourage a new $S^\prime$ less predictable to $Q$. By rearranging the equation such that the left hand side remains positive, we have arrived at a lower bound on $I(S^\prime : A | S=\bs{s}; \psi^{(i+1)}, \varphi^{(k+1)})$ which recovers the expected devaluation progress.

Equation \ref{eq:argneq} summarises the argument associated with Equation \ref{eq:mutinfo-relation} and \ref{eq:pardyn}. 
	\begin{equation}
	\begin{aligned} \label{eq:argneq}
	\varphi^{(k+1)} 
	&= 
	\argmax_{\varphi^{(k)}} 
	\Bigl[
	I(S^\prime : A|S=\bs{s}; \psi^{(i)}, \varphi^{(k)}) -
	\min_{\bar{\psi}^{(i)}} I(S^\prime : A|S=\bs{s}; \bar{\psi}^{(i)}, \varphi^{(k)})
	\Bigr] \\
	&\neq
	\argmax_{\varphi^{(k)}} 
	\Bigl[
	\min_{\psi^{(i)}} I(S^\prime : A|S=\bs{s}; \psi^{(i)}, \varphi^{(k)})
	\Bigr]
	\end{aligned}
	\end{equation}

Finally, we offer an intuition on how policy optimisation gives rise to heterostatic motivation. This is made clear from the optimised target $I(S^\prime : A | S=\bs{s}; \psi^{(i+1)}, \varphi^{(k+1)})$, found on the right hand side of Equation \ref{eq:mutinfo-relation}. It is instructive to re-introduce the true marginalisation $P(S^\prime|S=\bs{s}; \theta, \varphi)$ from Equation \ref{eq:mm}; write:
	\begin{equation}
	\begin{aligned} \label{eq:hetero}
	&\phantom{=..}
	I(S^\prime : A|S=\bs{s}; \psi^{(i+1)}, \varphi^{(k+1)}) \\
	&=
	\sum_{\bs{a}} \pi(\bs{a}|\bs{s}; \varphi^{(k+1)}) 
	\sum_{\bs{s}^\prime} P(\bs{s}^\prime|\bs{s}, \bs{a}; \theta) 
	\log\frac{P(\bs{s}^\prime|\bs{a}, \bs{s}; \theta)}{Q(\bs{s}^\prime|\bs{s}; \psi^{(i+1)})} \\
	&=
	\sum_{\bs{a}} \pi(\bs{a}|\bs{s}; \varphi^{(k+1)}) 
	\sum_{\bs{s}^\prime} P(\bs{s}^\prime|\bs{s}, \bs{a}; \theta) 
	\log
	\frac{P(\bs{s}^\prime|\bs{a}, \bs{s}; \theta)}  {P(\bs{s}^\prime|\bs{s}; \theta, \varphi^{(k+1)})}
	\frac{P(\bs{s}^\prime|\bs{s}; \theta, \varphi^{(k+1)})}  {Q(\bs{s}^\prime|\bs{s}; \psi^{(i+1)})} \\
	&=
	I(S^\prime : A|S=\bs{s}; \varphi^{(k+1)}) + 
	D_{KL} \bigr[ 
	P(\bs{s}^\prime|\bs{s}; \theta, \varphi^{(k+1)}) 
	\Vert 
	Q(\bs{s}^\prime|\bs{s}; \psi^{(i+1)}) \bigl]
	\end{aligned}
	\end{equation}
Simply, the optimised policy is such that the agent increases the conditional mutual information and is pushed away (via increasing the KL-divergence) from its homeostatic state $Q$.

\section{Implementation Considerations} \label{sec:impc}

This section presents practical considerations when motivating the aforementioned agent using neural networks. These considerations were mainly for the ease of calculating KL-divergence analytically. 

\subsection{Forward model}

We assumed, at the any given time, the state follows some Gaussian distribution with mean $\bs{s}$ and covariance $\Sigma$. The future state is described by its mean $\bs{s}^\prime$ according to the deterministic mapping $\bs{s}^\prime = f(\bs{a}, \bs{s}; \theta)$, where $\bs{a}$ is the action sampled from policy. $f$ represents a neural network with trainable parameter $\theta$:
	\begin{equation}
	\begin{aligned} \label{eq:imp-fm}
	f(\bs{a}, \bs{s}; \theta) &= \bs{A} \bs{s} + 
	\left( \sum_\iota a_{\iota} \bs{B}^{\iota} \right) \bs{s} + 
	\bs{C} \bs{a} + o
	\end{aligned}
	\end{equation}
$\bs{A}$, $\bs{B}$, and $\bs{C}$ are approximations of Jacobian matrices and $o$ a constant, all depending on $\theta$. $\bs{B}$ is a three-way tensor indexed by $\iota$ along the first axis. This treatment is similar to \citet{e2c} (also cf. \citet{dvbf}), except that we considered a bilinear approximation and that, in the following sections, we used only the mean states in a deterministic environment.

The above formalism follows that $\bs{s}^\prime$ has covariance matrix $\mathbb{E}[\bs{s}\bs{s}^{\prime\intercal}] = \bs{J} \bs{\Sigma} \bs{J}^\intercal$, where $\bs{J} = \left( \bs{A} + \sum_\iota a_\iota \bs{B}^\iota \right)$. The transition probability is then given by 
	\begin{equation}
	\begin{aligned} \label{eq:imp-fmcov}
	P(S^\prime | A=\bs{a}, S=\bs{s}; \theta) &= N \left(f(\bs{a}, \bs{s}; \theta), \bs{J} \bs{\Sigma} \bs{J}^\intercal \right)
	\end{aligned}
	\end{equation}

\subsection{Meta model}

Our meta model was defined as $Q(S^\prime|S=\bs{s}; \psi) = N(\bs{\mu}^\prime, \bs{\Sigma}^\prime; \psi)$, where mean $\bs{\mu}^\prime$ and covariance matrix $\bs{\Sigma}^\prime$ are outputs of a neural network parametrised by $\psi$. Specifically, the covariance matrix is constructed as follows:
	\begin{equation}
	\begin{aligned} \label{eq:imp-mmcov}
	\bs{\Sigma}^\prime &= \bs{H} \bs{D} \bs{H}^\intercal, \;\; \bs{D} = \mathrm{diag}(\bs{d}) \\
	\bs{H} &= \bs{I} - 2\frac{\bs{vv}^\intercal}{\norm{\bs{v}}^2},
	\end{aligned}
	\end{equation}
where $\bs{d}$ is a positive-valued vector, $\bs{I}$ an identity matrix, and $\bs{v}$ a Householder vector \citep{householderflow}.


Whenever possible, e.g., employing Experience Replay, gradients of the objective may be weighted by the probability ratio $P(\bs{s}^\prime|\bs{a}, \bs{s}; \theta^{(\ell+1)}) / {P(\bs{s}^\prime|\bs{a}, \bs{s}; \theta^{(\ell)})}$, where the superscripts $(\ell+1)$ and $(\ell)$ indicate forward model parameters after and before gradient updates. This procedure encourages boredom to be properly induced in accordance with forward model learning.

\section{Experiment}

To verify whether our algorithm exhibits online curiosity, we focused on benchmarking agent's forward model under these constraints: i) agent should learn to bootstrap its own training set; ii) the probability of visiting different states is not uniformly distributed; iii) the amount of time to accumulate training data points is limited. 

Boredom-based benchmarks were compared 1) against (ideal) models trained using oracle dataset, and 2) with reduced models under the pruning hierarchy (Section \ref{subsec:mod-prune}, Table \ref{tab:model-prune}).

\subsection{Training environment}

\begin{figure}[H]
		\begin{center}
			\includegraphics[width=8cm]{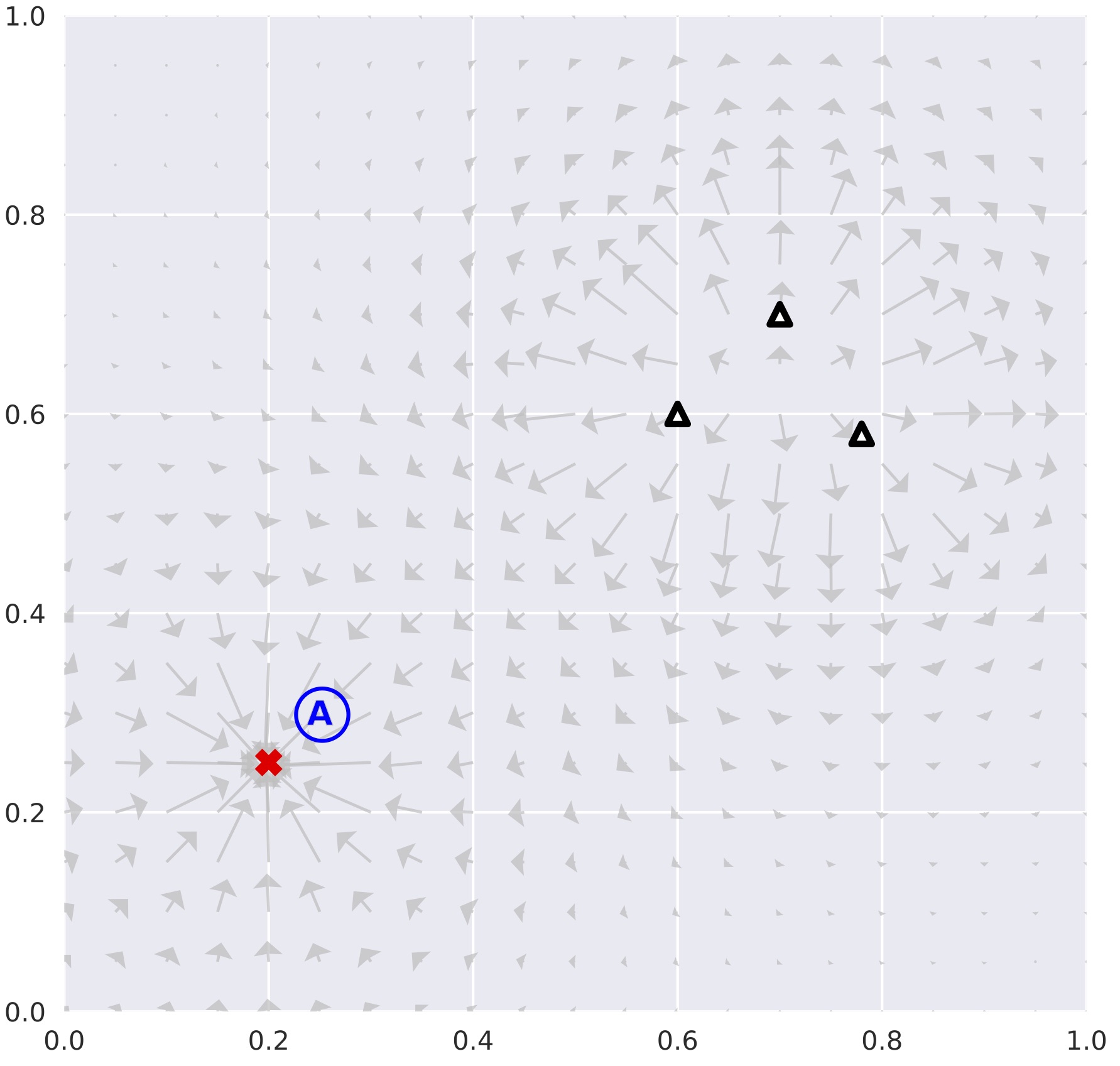}
		\end{center}
		\caption{Environmental configuration. The red cross represents attractor, whilst black triangles repellers. Vector plots indicate the forces exerted if the agent assumed the positions with zero velocities. The initial position is set at the blue letter `A'. This configuration remains identical cross all model variants and test runs.}
		\label{fig:env}
\end{figure}

Our agents were tested in a physics simulator, free of stochasticity, built to expand the classical Mountain Car environment (e.g., `MountainCar-v0' included in \citet{aigym}) into two-dimensional state space. The environment is analogous to the Mountain Car in ways that it has attractors and repellers that resemble hill- and valley-like landscapes (Figure \ref{fig:env}). The presence of both structures serves as acceleration modifier to the agent. This makes state visitation biased toward attractors. Therefore, the acquisition of an accurate forward model necessitates planning visits to the vicinity of repellers.

The states an agent can occupy were defined as the tuple $(x, y, \dot x, \dot y)$ in continuous real space. Positions $(x, y) \in [0, 1]^2$ were bounded in a unit square, whereas velocities $(\dot x, \dot y)$ were not. Boundary condition resets $x$ and $y$ to zero velocities. However, it is possible for the agent to slide along the boundaries if its action goes in the direction parallel to the nearby boundary. We note that being trapped in the corners is possible; though an agent could potentially get itself unstuck if appropriate actions were carried out. 

Agent's action policy was represented by a categorical distribution over accelerations in $x$ and $y$ directions. The distribution was defined on the interval $[-2.0, 2.0]^2$, evenly divided into a $11\times 11$ grid. When an action is selected, the corresponding acceleration is modified according to forces exerted by the attractors and repellers.

Unlike the classical Mountain Car, our environment does not express external rewards, nor does it possess any states that are indicative of termination. Agents were allowed a pre-defined time limit ($T=30,000$ steps; {\it Data Accumulation Phase} or DAP) to act without interruption. Agent's experiences in terms of state transitions were collected in a database, which was sampled from for training at each step. During DAP, learning rates for model parameters remained constant. After DAP (or {\it post}-DAP), agent entered an action-free stage lasted for $T=30,000$, during which only sampling from own experience pool for forward model training was performed. Learning rate scheduling scheme was implemented at post-DAP.

\subsection{Oracle dataset}

To contrast with self-assisted data accumulation, we constructed an oracle dataset, with which a (forward) model was trained. We referred to this class of model as Oracle. The oracle dataset assumed unbiased state occupancy and action choice. Specifically, we acquired the dataset by evenly dividing the state-action space into a $49 \times 49 \times 11 \times 11 \times 11 \times 11$ grid. Each state-action pair was passed to the physics simulator to evaluate the future state. The oracle dataset differs from self-assisted ones in that contained positions near the repellers that an agent is incapable of visiting.

The training, testing, and validation sets were prepared by re-sampling the resulting dataset without replacement according to the ratio $0.8$, $0.16$, and $0.04$. The model was trained for $60,000$ epochs. During training, the learning rate was scheduled according to test error. Benchmarking was performed on the validation set as part of model comparisons (see Section \ref{subsec:modcompare}).

\subsection{Model pruning} \label{subsec:mod-prune}

We defined five variants of our boredom-driven curious agent. With each variation, the agent receives cumulative reductions in network components. Theses reductions are summarised as model pruning hierarchy in Table \ref{tab:model-prune}. 

The reason that we motivated model comparisons based on model pruning is as follows. Overall, as model pruning progresses the agent was deprived of connections with constructs like devaluation progress, intrinsic motivation, and planning. Eventually, the agent lost the ability to contextualise action selection and became a random-walk object. At the stage, the agent was still explorative in a crude sense, albeit not in a curious way. Through these treatments we demonstrated the impact boredom and intrinsic motivation have on model learning.

\subsubsection{Boredom-driven curiosity (C/B)}

The first agent variant retained all distinctive components introduced in Section \ref{sec:algorithm}. The meta-model provides the devaluation progress as intrinsic rewards, whilst the value function enables the agent to plan actions that are intrinsically rewarding in the long run.

\subsubsection{Predictive error-driven curiosity (C/PE)}

The C/PE variant tests whether the induction of boredom is a constructive form of intrinsic motivation. This is achieved by removing the meta-model, thereby requiring an alternative definition of intrinsic reward. We replaced the devaluation progress with {\it learning progress} defined by mean squared errors of the forward model:
	\begin{equation}
	\begin{aligned}
	R_\theta^{(\ell + 1)} 
	&:= 
	\mathcal{L}_{fm}(\theta^{(\ell)}) - \mathcal{L}_{fm}(\theta^{(\ell+1)}) \\
	\mathcal{L}_{fm}(\theta)
	&:= 
	\mathcal{L}(\bs{s}^\prime, \bs{a}, \bs{s}; \theta) \\
	&\phantom{:}=
	\Vert \bs{s}^\prime - f(\bs{a}, \bs{s}; \theta) \Vert^2
	\end{aligned}
	\end{equation}
The construction of learning progress is one typical approach to intrinsic motivation and curiosity \citep{pathak2017icm, schmidhuber1991}.

\subsubsection{Policy gradients, intrinsic reward samples (PG/IRS), Gaussian rewards (PG/GR)}

Next, we examined how reward statistics alone influences policy update and, as a consequence, model learning. The value function was removed at this stage to dissociate policy learning from any downstream effects of value learning. 

One distinctive feature of devaluation progress is that it entails time-varying rewards --- depending on the amount of time over which an agent has evolved in the environment. We hypothesised that the emergence of curious policy is associated with reward dynamics over time. That is to say, if one perturbs the magnitudes and directions of the policy gradients with reward statistics appropriate for the ongoing time frame, the agent should exhibit similar curious behaviours. Nevertheless, we argue that such treatment is only sensible given virtually identical initial conditions. Specifically, all agent variants shared the same, environmental configuration, initial position, and network initialisation.

To this end, we prepared a database for intrinsic reward samples. During C/B performance, all reward samples were collected and labelled with the corresponding time step. Afterwards, the PG/IRS agents randomly sampled from the database in a temporally synchronised manner and applied standard policy gradients.

The PG/IRS was contrasted with the PG/GR variant. Their difference lies in that a surrogate reward was used in place of the database. We defined the surrogate reward as a Gaussian distribution with time-invariant parameters, in which the mean $\mu=0$ is under the assumption of equilibrium devaluation progress and the standard deviation $\sigma=0.01$, as derived from the entire database.

\subsubsection{Random-walk policy (P/RW)}

Finally, we constructed a random-walk agent. All network components, apart from the forward model, were removed. This agent variant represents the case without intrinsic motivation and is agnostic to curiosity. Broadly speaking, the agent was still explorative due to its maximum entropy action policy. We regarded this version as the worse case scenario to contrast with the rest of the variants.

\subsection{Model comparisons} \label{subsec:modcompare}

All model variants were compared on the basis of validation error given the oracle dataset. We performed 128 runs for each of the six variants (Oracle, C/B, C/PE, PG/IRS, PG/GR, and P/RW). All variants, across all runs, were assigned to identical environmental configuration (e.g., initial position, attractor/repeller placements). Network components, whenever applicable, shared identical architecture and were trained with consistent batch size and learning rate. Model parameters followed the Xavier initialisation \citep{glorot}. During post-DAP, learning rate scheduling was implemented such that a factor $0.1$ reduction was applied upon a 3000-epoch loss plateau.

\section{Results}

We first characterised individual agent variants' qualities of being i) explorative and ii) perseverative. Active exploration is one defining attribute of curiosity \citep{2013gottlieb}, simply because it differentiates between uncertain and known situations, thus giving rise to effective information acquisition. This, however, should be complemented with bounded perseverance; namely, to prevent oneself from being permanently or dynamically captured---i.e., by the corners or the attractor. 

\begin{figure}[H]
		\begin{center}
			\includegraphics[width=15cm]{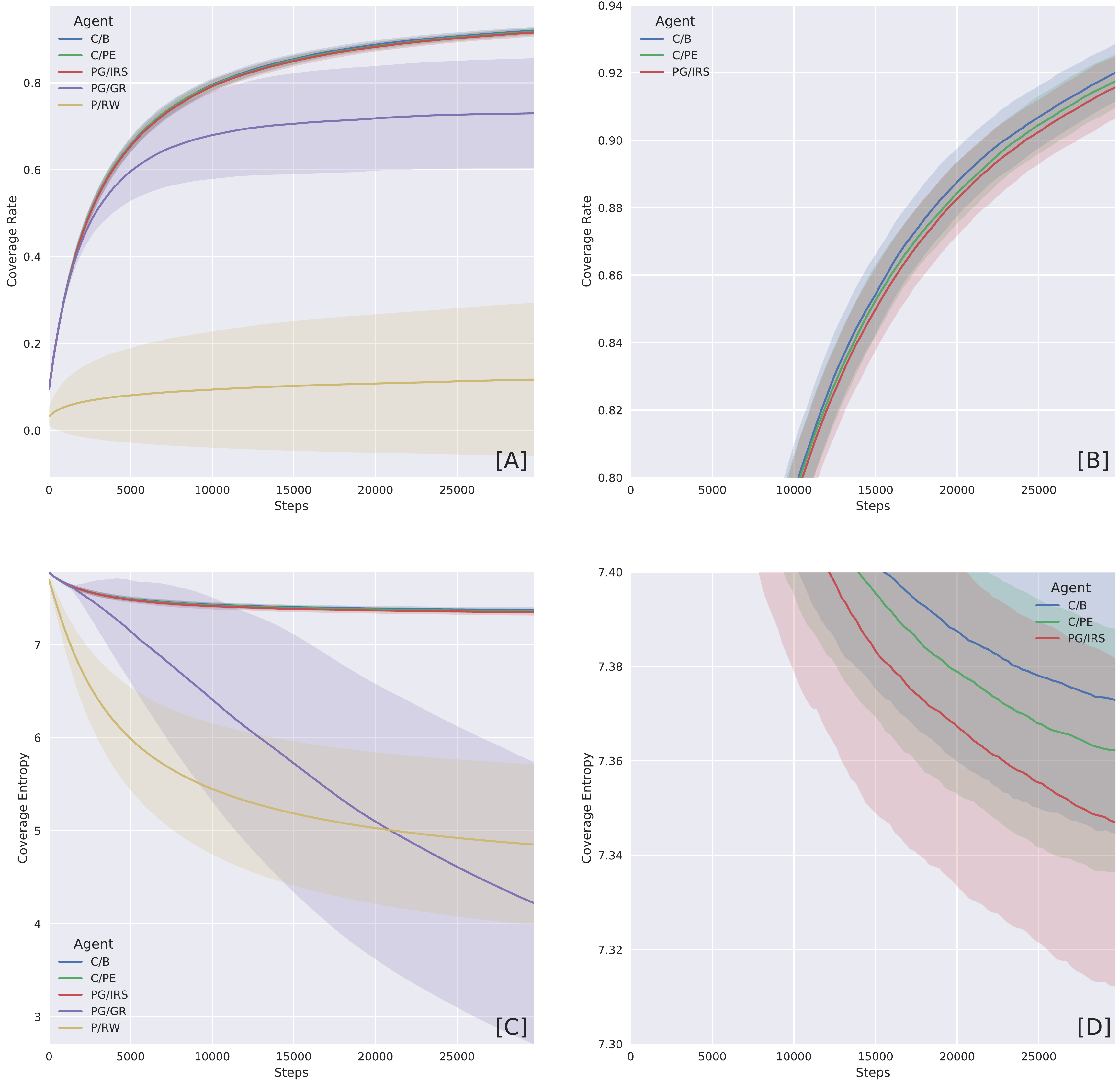}
		\end{center}
		\caption{Coverage Rate (CR) and Coverage Entropy (CE) by agent variants. The two measures were computed by first turning the state space into a $50\times 50$ grid, ignoring velocities. CR then marks over time whether or not a cell has been visited. Whereas, CE treats the grid as a probability distribution. Starting with maximum entropy, CR cumulatively counts the number of times a position is being visited. Entropy was calculated at each time step using normalised counter. [A] Overview of CR shows the distinction between curious and non-curious agents. Curiosity caused the agents to explore faster. [B] Close-up on the curious agent variants, which were equally explorative. [C] Overview of CE shows agents with different levels of perseverance. The P/RW variants were captured by the attractor, whilst the PG/GR variants were prone to blockage. [D] Close-up on curious agents, which were characterised by higher CE due to attractor avoidance and more frequent repeller visitation attempts. Shaded regions represent one standard deviation.}
		\label{fig:covent}
\end{figure}

The two qualities can be distinguished, as shown in Figure \ref{fig:covent}, by respective measures of Coverage Rate (CR) and Coverage Entropy (CE). The two measures were computed by first turning the state space into a $50\times 50$ grid, ignoring velocities. CR then marks over time whether or not a cell has been visited. Whereas, CE treats the grid as a probability distribution. Starting with maximum entropy, CR cumulatively counts the number of times a position is being visited. Entropy was calculated at each time step using normalised counter.

Because (state) visitation bias was inherent in our testing environment, naturally, agents occupying a subset of states would cause CE to reduce faster than those who attempted to escape. The C/B, C/PE, and PG/IRS variants were regarded as curious and intrinsically motivated. Our results showed that these variants were predominantly explorative and non-perseverative. By contrast, the P/RW agent, albeit explorative, had no principled means to escape the dynamic lock. However, if $t\to\infty$ the P/RW should be able to explore further by chance. The PG/GR variant, on the other hand, exhibited, intermediate explorativeness and extreme perseverance with disproportionately high variance. We attributed this behaviour to the detrimental effects of inappropriately informative reward statistics.

\begin{figure}[H]
	\begin{center}
	\includegraphics[width=15cm]{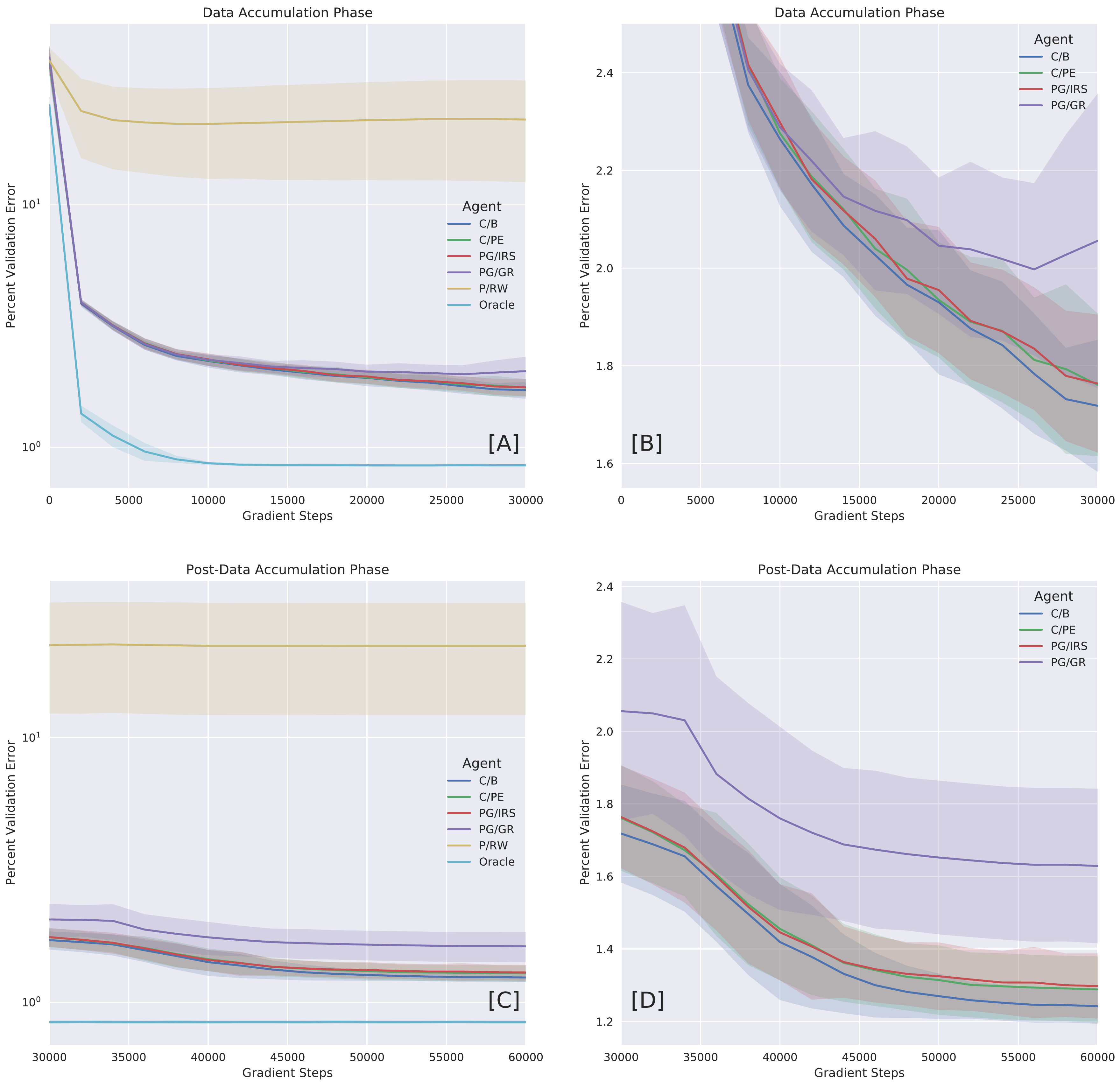}
	\end{center}
	\caption{Benchmarking model variants with oracle dataset. Performances were reported in error percentage (also, see Table \ref{tab:vldprogress}). [A] Performance as a function of time during Data Accumulation Phase (DAP). [B] Close-up on curious variants (C/B, C/PE, and PG/IRS), as well as policy gradients (PG/GR) informed by surrogate reward statistics. The C/PE and PG/IRS variants performed similarly, but differed significantly from C/B (Table \ref{tab:stats}). [C] Performance over time during post-DAP. [D] Close-up on post-DAP performances for curious variants and PG/GR.}
	\label{fig:vldprogress}
\end{figure}

Next, we benchmarked forward model performance of individual variants by their validation loss and error percentage. We reported DAP and post-DAP performances separately as a function of time in Figure \ref{fig:vldprogress}. Error percentage was calculated as the percent ratio between root mean squared loss and the maximum pair-wise Euclidean distance in the validation set.

The Oracle model, trained under the supervision of oracle training set, reached an error percentage of $0.84\%$ for both DAP and post-DAP, amounting to approximately 30\% improvement over the terminal performance of C/B variant. All variants considered curious (C/B, C/PE, and PG/IRS) had similar performances during DAP. In particular, the PG/IRS, which received independent intervention from the `true' reward distributions achieved marginally lower performance but indistinguishable from the C/PE variant. This outcome was observed for both DAP and post-DAP, suggesting intrinsic reward samples derived from C/B contributed favourably even to the standard policy gradients algorithm. 

Though without the ability to approximate value function, the PG/IRS variant underperformed in benchmarking, as compared with the value-enabled, C/B variant. Using non-parametric test, the difference was detected for DAP ($p=$ 0.0006) and post-DAP ($p=$ 6.4E-8), respectively. Similar observations were also made for comparisons between C/B and C/PE, at $p=$ 0.0029 (DAP) and $p=$ 5.9E-5 (post-DAP). Overall, this suggested significant differences in the experiences accumulated across agent variants. The aforementioned statistics were reported in Table \ref{tab:vldprogress} and \ref{tab:stats}.

\section{Conclusion}

We have provided a formal account on the emergence of boredom from an information-seeking perspective and addressed its constructive role in enabling curious behaviour. We envisaged actions as instrumental in agent's epistemic disclosure, which is assimilated by another conditionally independent cognitive process. This led to the central claim of this study---pertaining to superior data-gathering efficiency and hence effective curiosity. We supported this claim with empirical evidence, showing that boredom-enabled agents consistently outperformed other curious agents in self-assisted world model learning. Our results solicited the interpretation that the relationship between homeostatic and heterostatic intrinsic motivations can in fact be complementary; therefore, we have offered one unifying perspective for the intrinsic motivation landscape.

\section*{Conflict of Interest Statement}

The authors declare that the research was conducted in the absence of any commercial or financial relationships that could be construed as a potential conflict of interest. All authors were employed by Araya, Inc.

\section*{Author Contributions}
YY conceived of this study, performed the experiments, and wrote the first draft of the manuscript. AYCC programmed the physics simulator, wrote part of Introduction, and created Figure \ref{fig:intuition}. All authors contributed to manuscript revision, read and approved the submitted version.

\section*{Funding}
This study was funded by the Japan Science and Technology Agency (JST) under CREST grant number JPMJCR15E2.

\section*{Acknowledgments}
YY would like to thank Martin Biehl and Ildefons Magrans de Abril for insightful discussions.

\bibliographystyle{plainnat}
\bibliography{ms}

\begin{thebibliography}{32}
\providecommand{\natexlab}[1]{#1}
\providecommand{\url}[1]{\texttt{#1}}
\expandafter\ifx\csname urlstyle\endcsname\relax
  \providecommand{\doi}[1]{doi: #1}\else
  \providecommand{\doi}{doi: \begingroup \urlstyle{rm}\Url}\fi

\bibitem[Adams(1982)]{adams1982variations}
Christopher~D Adams.
\newblock Variations in the sensitivity of instrumental responding to
  reinforcer devaluation.
\newblock \emph{The Quarterly Journal of Experimental Psychology}, 34\penalty0
  (2):\penalty0 77--98, 1982.

\bibitem[Adams and Dickinson(1981)]{adams1981instrumental}
Christopher~D Adams and Anthony Dickinson.
\newblock Instrumental responding following reinforcer devaluation.
\newblock \emph{The Quarterly Journal of Experimental Psychology Section B},
  33\penalty0 (2b):\penalty0 109--121, 1981.

\bibitem[Balleine and Dickinson(1998)]{Balleine.1998}
Bernard~W. Balleine and Anthony Dickinson.
\newblock Goal-directed instrumental action: Contingency and incentive learning
  and their cortical substrates.
\newblock \emph{Neuropharmacology}, 37\penalty0 (4-5):\penalty0 407--419, 1998.
\newblock ISSN 00283908.
\newblock \doi{10.1016/S0028-3908(98)00033-1}.

\bibitem[Bellemare et~al.(2016)Bellemare, Srinivasan, Ostrovski, Schaul,
  Saxton, and Munos]{bellemare2016count}
Marc Bellemare, Sriram Srinivasan, Georg Ostrovski, Tom Schaul, David Saxton,
  and Remi Munos.
\newblock Unifying count-based exploration and intrinsic motivation.
\newblock In \emph{Advances in Neural Information Processing Systems}, pages
  1471--1479, 2016.

\bibitem[Bench and Lench(2013)]{bench2013function}
Shane~W Bench and Heather~C Lench.
\newblock On the function of boredom.
\newblock \emph{Behavioral Sciences}, 3\penalty0 (3):\penalty0 459--472, 2013.

\bibitem[{Brockman} et~al.(2016){Brockman}, {Cheung}, {Pettersson},
  {Schneider}, {Schulman}, {Tang}, and {Zaremba}]{aigym}
G.~{Brockman}, V.~{Cheung}, L.~{Pettersson}, J.~{Schneider}, J.~{Schulman},
  J.~{Tang}, and W.~{Zaremba}.
\newblock {OpenAI Gym}.
\newblock \emph{ArXiv e-prints}, June 2016.

\bibitem[Fahlman et~al.(2009)Fahlman, Mercer, Gaskovski, Eastwood, and
  Eastwood]{fahlman2009does}
Shelley~A Fahlman, Kimberley~B Mercer, Peter Gaskovski, Adrienne~E Eastwood,
  and John~D Eastwood.
\newblock Does a lack of life meaning cause boredom? results from psychometric,
  longitudinal, and experimental analyses.
\newblock \emph{Journal of social and clinical psychology}, 28\penalty0
  (3):\penalty0 307--340, 2009.

\bibitem[Fairbank and Alonso(2012)]{2012fairbank}
Michael Fairbank and Eduardo Alonso.
\newblock Value-gradient learning.
\newblock In \emph{Neural Networks (IJCNN), The 2012 International Joint
  Conference on}, pages 1--8. IEEE, 2012.

\bibitem[Friston et~al.(2012)Friston, Thornton, and Clark]{darkroom}
Karl Friston, Christopher Thornton, and Andy Clark.
\newblock Free-energy minimization and the dark-room problem.
\newblock \emph{Frontiers in psychology}, 3:\penalty0 130, 2012.

\bibitem[Friston et~al.(2017)Friston, Lin, Frith, Pezzulo, Hobson, and
  Ondobaka]{friston2017curiosity}
Karl~J Friston, Marco Lin, Christopher~D Frith, Giovanni Pezzulo, J~Allan
  Hobson, and Sasha Ondobaka.
\newblock Active inference, curiosity and insight.
\newblock \emph{Neural computation}, 29\penalty0 (10):\penalty0 2633--2683,
  2017.

\bibitem[Glorot and Bengio(2010)]{glorot}
Xavier Glorot and Yoshua Bengio.
\newblock Understanding the difficulty of training deep feedforward neural
  networks.
\newblock In \emph{Proceedings of the thirteenth international conference on
  artificial intelligence and statistics}, pages 249--256, 2010.

\bibitem[Gottlieb et~al.(2013)Gottlieb, Oudeyer, Lopes, and
  Baranes]{2013gottlieb}
Jacqueline Gottlieb, Pierre-Yves Oudeyer, Manuel Lopes, and Adrien Baranes.
\newblock Information-seeking, curiosity, and attention: computational and
  neural mechanisms.
\newblock \emph{Trends in cognitive sciences}, 17\penalty0 (11):\penalty0
  585--593, 2013.

\bibitem[Harris(2000)]{harris2000correlates}
Mary~B Harris.
\newblock Correlates and characteristics of boredom proneness and boredom.
\newblock \emph{Journal of Applied Social Psychology}, 30\penalty0
  (3):\penalty0 576--598, 2000.

\bibitem[{Heess} et~al.(2015){Heess}, {Wayne}, {Silver}, {Lillicrap}, {Tassa},
  and {Erez}]{svg}
N.~{Heess}, G.~{Wayne}, D.~{Silver}, T.~{Lillicrap}, Y.~{Tassa}, and T.~{Erez}.
\newblock {Learning Continuous Control Policies by Stochastic Value Gradients}.
\newblock \emph{ArXiv e-prints}, October 2015.

\bibitem[{Karl} et~al.(2016){Karl}, {Soelch}, {Bayer}, and {van der
  Smagt}]{dvbf}
M.~{Karl}, M.~{Soelch}, J.~{Bayer}, and P.~{van der Smagt}.
\newblock {Deep Variational Bayes Filters: Unsupervised Learning of State Space
  Models from Raw Data}.
\newblock \emph{ArXiv e-prints}, May 2016.

\bibitem[{Lillicrap} et~al.(2015){Lillicrap}, {Hunt}, {Pritzel}, {Heess},
  {Erez}, {Tassa}, {Silver}, and {Wierstra}]{deepmind2015lillicrap}
T.~P. {Lillicrap}, J.~J. {Hunt}, A.~{Pritzel}, N.~{Heess}, T.~{Erez},
  Y.~{Tassa}, D.~{Silver}, and D.~{Wierstra}.
\newblock {Continuous control with deep reinforcement learning}.
\newblock \emph{ArXiv e-prints}, September 2015.

\bibitem[London et~al.(1972)London, Schubert, and Washburn]{london1972increase}
Harvey London, Daniel~S Schubert, and Daniel Washburn.
\newblock Increase of autonomic arousal by boredom.
\newblock \emph{Journal of Abnormal Psychology}, 80\penalty0 (1):\penalty0 29,
  1972.

\bibitem[Mannella et~al.(2016)Mannella, Mirolli, and
  Baldassarre]{mannella2016goal}
Francesco Mannella, Marco Mirolli, and Gianluca Baldassarre.
\newblock Goal-directed behavior and instrumental devaluation: A neural
  system-level computational model.
\newblock \emph{Frontiers in behavioral neuroscience}, 10:\penalty0 181, 2016.

\bibitem[{Mnih} et~al.(2013){Mnih}, {Kavukcuoglu}, {Silver}, {Graves},
  {Antonoglou}, {Wierstra}, and {Riedmiller}]{deepmind2013atari}
V.~{Mnih}, K.~{Kavukcuoglu}, D.~{Silver}, A.~{Graves}, I.~{Antonoglou},
  D.~{Wierstra}, and M.~{Riedmiller}.
\newblock {Playing Atari with Deep Reinforcement Learning}.
\newblock \emph{ArXiv e-prints}, December 2013.

\bibitem[Ostrovski et~al.(2017)Ostrovski, Bellemare, Oord, and
  Munos]{ostrovski2017count}
Georg Ostrovski, Marc~G Bellemare, Aaron van~den Oord, and R{\'e}mi Munos.
\newblock Count-based exploration with neural density models.
\newblock \emph{arXiv preprint arXiv:1703.01310}, 2017.

\bibitem[Oudeyer and Kaplan(2009)]{intrinsicmotiv}
Pierre-Yves Oudeyer and Frederic Kaplan.
\newblock What is intrinsic motivation? a typology of computational approaches.
\newblock \emph{Frontiers in neurorobotics}, 1:\penalty0 6, 2009.

\bibitem[Pathak et~al.(2017)Pathak, Agrawal, Efros, and Darrell]{pathak2017icm}
Deepak Pathak, Pulkit Agrawal, Alexei~A. Efros, and Trevor Darrell.
\newblock Curiosity-driven exploration by self-supervised prediction.
\newblock In \emph{ICML}, 2017.

\bibitem[Perkins and Hill(1985)]{perkins1985cognitive}
Rachel~E Perkins and AB~Hill.
\newblock Cognitive and affective aspects of boredom.
\newblock \emph{British Journal of Psychology}, 76\penalty0 (2):\penalty0
  221--234, 1985.

\bibitem[Schmidhuber(1991)]{schmidhuber1991}
J{\"u}rgen Schmidhuber.
\newblock A possibility for implementing curiosity and boredom in
  model-building neural controllers.
\newblock In \emph{Proc. of the international conference on simulation of
  adaptive behavior: From animals to animats}, pages 222--227, 1991.

\bibitem[Schmidhuber(2008)]{schmidhuber2008driven}
J{\"u}rgen Schmidhuber.
\newblock Driven by compression progress: A simple principle explains essential
  aspects of subjective beauty, novelty, surprise, interestingness, attention,
  curiosity, creativity, art, science, music, jokes.
\newblock In \emph{Workshop on Anticipatory Behavior in Adaptive Learning
  Systems}, pages 48--76. Springer, 2008.

\bibitem[Schubert(1977)]{schubert1977boredom}
Daniel~SP Schubert.
\newblock Boredom as an antagonist of creativity.
\newblock \emph{The Journal of Creative Behavior}, 11\penalty0 (4):\penalty0
  233--240, 1977.

\bibitem[Schubert(1978)]{schubert1978creativity}
Daniel~SP Schubert.
\newblock Creativity and coping with boredom.
\newblock \emph{Psychiatric Annals}, 8\penalty0 (3):\penalty0 46--54, 1978.

\bibitem[Sutton and Barto(1998)]{sutton1998reinforcement}
Richard~S Sutton and Andrew~G Barto.
\newblock \emph{Reinforcement learning: An introduction}, volume~1.
\newblock MIT press Cambridge, 1998.

\bibitem[{Tomczak} and {Welling}(2016)]{householderflow}
J.~M. {Tomczak} and M.~{Welling}.
\newblock {Improving Variational Auto-Encoders using Householder Flow}.
\newblock \emph{ArXiv e-prints}, November 2016.

\bibitem[van Tilburg and Igou(2012)]{van2012boredom}
Wijnand~AP van Tilburg and Eric~R Igou.
\newblock On boredom: Lack of challenge and meaning as distinct boredom
  experiences.
\newblock \emph{Motivation and Emotion}, 36\penalty0 (2):\penalty0 181--194,
  2012.

\bibitem[Vodanovich et~al.(1991)Vodanovich, Verner, and
  Gilbride]{vodanovich1991boredom}
Stephen~J Vodanovich, Kathryn~M Verner, and Thomas~V Gilbride.
\newblock Boredom proneness: Its relationship to positive and negative affect.
\newblock \emph{Psychological reports}, 69\penalty0 (3\_suppl):\penalty0
  1139--1146, 1991.

\bibitem[{Watter} et~al.(2015){Watter}, {Springenberg}, {Boedecker}, and
  {Riedmiller}]{e2c}
M.~{Watter}, J.~T. {Springenberg}, J.~{Boedecker}, and M.~{Riedmiller}.
\newblock {Embed to Control: A Locally Linear Latent Dynamics Model for Control
  from Raw Images}.
\newblock \emph{ArXiv e-prints}, June 2015.

\end{thebibliography}



\section*{Figure captions}

\section*{Tables}

\begin{table}[H]
	\begin{center}
	\caption{Model pruning hierarchy. Ticks mark the existence or dependence of trainable network component; circles indicate independent intervention. Top row: P/RW, random-walk policy; PG/GR, policy gradients with rewards drawn from Gaussian distribution; PG/IRS, policy gradients with intrinsic reward samples; C/PE, curiosity using forward model error; C/B, curiosity from boredom. First column: FM, forward model; AP, action policy;  IR, intrinsic rewards; VF, value function approximator; MM, meta-model.}
	\label{tab:model-prune}
	\begin{tabular}{r c c c c c c}
	\toprule
	& Oracle & P/RW & PG/GR & PG/IRS & C/PE & C/B \\ 
	\midrule
	FM & $\checkmark$ & $\checkmark$ & $\checkmark$ & $\checkmark$ & $\checkmark$ & $\checkmark$ \\ 
	AP &  & $\bigcirc$  & $\checkmark$ & $\checkmark$ & $\checkmark$ & $\checkmark$ \\ 
	IR &  &  &  & $\bigcirc$ & $\checkmark$ & $\checkmark$ \\ 
	VF &  &  &  &  & $\checkmark$ & $\checkmark$ \\ 
	MM &  &  &  &  &  & $\checkmark$ \\ 
	\bottomrule
	\end{tabular} 
	\end{center}
\end{table}

\begin{table}[H]
	\begin{center}
	\caption{Summary statistics on validation loss and error percentage as benchmarking scores. Key: DAP, Data Accumulation Phase; SD, standard deviation. For agent codes, see Table \ref{tab:model-prune}.}
	\label{tab:vldprogress}
	\begin{tabular}{r l p{2.3cm} | l p{2.3cm}}
	\toprule
	\multirow{2}{*}{Agent} & \multicolumn{2}{c}{DAP} & \multicolumn{2}{c}{Post-DAP} \\
	\cline{2-5} \\
	& MSE loss (SD) & Mean Percent Error (SD) & MSE loss (SD) & Mean Percent Error (SD) \\
	\midrule
	\multirow{2}{*}{Oracle}
			 & 0.0008   & 0.8430   & 0.0008   & 0.8428   \\
		     & (2.3E-5) & (0.0123) & (2.2E-5) & (0.0114) \\
	\multirow{2}{*}{C/B}
			 & 0.0033   & 1.7181   & 0.0017   & 1.2420   \\
			 & (0.0006) & (0.1357) & (0.0001) & (0.0488) \\
	\multirow{2}{*}{C/PE}
			 & 0.0035   & 1.7611   & 0.0019   & 1.2882   \\	
			 & (0.0006) & (0.1464) & (0.0003) & (0.0916) \\
	\multirow{2}{*}{PG/IRS}
			 & 0.0035   & 1.7637   & 0.0020   & 1.2976   \\
			 & (0.0006) & (0.1418) & (0.0003) & (0.0902) \\
	\multirow{2}{*}{PG/GR}
			 & 0.0048   & 2.0559   & 0.0030   & 1.6288   \\
			 & (0.0017) & (0.3026) & (0.0008) & (0.2140) \\
	\multirow{2}{*}{P/RW}
			 & 0.6663   & 22.2734   & 0.6615   & 22.1453   \\
			 & (0.3904) & (10.0085) & (0.3864) & (10.0775) \\
	\bottomrule
	\end{tabular}
	\end{center}
\end{table}

\begin{table}[H]
	\begin{center}
	\caption{Non-parametric statistical tests comparing terminal performance at DAP and post-DAP for curious model variants.}
	\label{tab:stats}
	\begin{tabular}{c c c c}
	\multicolumn{4}{c}{Mann-Whitney U Test ($n=128, \alpha=0.025$, Bonferroni corrected)} \\
	\toprule
	\multicolumn{2}{c}{Validation loss}
	& DAP ($T=30000$) & Post-DAP ($T=60000$) \\
	\midrule
	\multirow{2}{*}{C/B $<$ C/PE} 
		& Statistics    & 6558.0 & 5911.0 \\
		& {\it p}-value & 0.0029 & 5.9E-5 \\ 
	\multirow{2}{*}{C/B $<$ PG/IRS} 
		& Statistics    & 6275.0 & 5062.0 \\
		& {\it p}-value & 0.0006 & 6.4E-8 \\
	\bottomrule
	\end{tabular}
	\end{center}
\end{table}


\section*{Algorithms}

\begin{algorithm}[H]
	\caption{Homeo-heterostatic value gradients}
	\label{alg:hhvg}
	\begin{algorithmic}[1]
		\State \textbf{Variables} \par
		\Statex \hskip\algorithmicindent\hskip\algorithmicindent 
			outer loop time $t$ \par
		\Statex \hskip\algorithmicindent\hskip\algorithmicindent 
			gradient step counter $\ell, i, j, k$ \par
		\Statex \hskip\algorithmicindent\hskip\algorithmicindent 
			state $\bs{s}^t := \bs{s}(t)$ and action $\bs{a}^t := \bs{a}(t)$ \par
		\Statex \hskip\algorithmicindent\hskip\algorithmicindent 
			learning rate $\lambda^\theta, \lambda^\psi, \lambda^\nu, \lambda^\varphi$ \par 
		\Statex \hskip\algorithmicindent\hskip\algorithmicindent discount factor 
			$\gamma$ \par 
		\Statex \hskip\algorithmicindent\hskip\algorithmicindent 
			experience pool $\mathcal D$

		\State \textbf{Models and parameters} \par 
		\Statex \hskip\algorithmicindent\hskip\algorithmicindent 
			forward model $P(S^\prime|\bs{s}, \bs{a}; \theta)$ \par
		\Statex \hskip\algorithmicindent\hskip\algorithmicindent 
			meta-model $Q(S^\prime|\bs{s};\psi)$ \par
		\Statex \hskip\algorithmicindent\hskip\algorithmicindent 
			value apprximator $\hat{V}(\bs{s}; \nu)$ \par
		\Statex \hskip\algorithmicindent\hskip\algorithmicindent 
			action policy $\pi(A|\bs{s}; \varphi)$

		\State \textbf{Objectives} \par
		\Statex \hskip\algorithmicindent\hskip\algorithmicindent
			forward-model learning $\mathcal{L}_{fm}(\theta)$ \par
		\Statex \hskip\algorithmicindent\hskip\algorithmicindent
			meta-model learning $\mathcal{L}_{mm}(\psi)$ \Comment{Eq.\ref{eq:mm-loss}} \par
		\Statex \hskip\algorithmicindent\hskip\algorithmicindent
			value learning $\mathcal{L}_{vf}(\nu)$ \Comment{Eq.\ref{eq:vf-loss}} \par
		\Statex \hskip\algorithmicindent\hskip\algorithmicindent
			policy learning $\mathcal{L}_{ap}(\varphi)$ \Comment{Eq.\ref{eq:policy-loss}} \par
		
		\For {$t=0\dots T$}
		\State From $\bs{s}^t$, sample action $\bs{a}^t \sim \pi(\cdot | \bs{s}^t;\varphi)$
		\State Perform $\bs{a}^t$ and advance to $\bs{s}^{t+1}$
		\State Insert tuple $\left( \bs{s}^t, \bs{a}^t, \pi(\bs{a}^t|\bs{s}^t), \bs{s}^{t+1} \right)$ into $\mathcal D$
		
		\State Sample $\mathcal D$ and train forward model: \par
		\State \hskip\algorithmicindent 
			$\mathcal{L}_{fm}(\theta) := 
			\mathcal{L}(\bs{s}^\prime, \bs{a}, \bs{s}; \theta) = 
			\Vert \bs{s}^\prime - f(\bs{a}, \bs{s}; \theta) \Vert^2$ \Comment{Eq.\ref{eq:imp-fm}}\par
		\State \hskip\algorithmicindent 
			$\theta^{(\ell+1)} \gets \theta^{(\ell)} - 
			\lambda_\theta \nabla_\theta \mathcal{L}_{fm}(\theta^{(\ell)})$ \par
		
		\State Value learning ($M$ updates, see Algorithm \ref{alg:fpeval})
		
		\State Sample $\mathcal D$ and perform devaluation: \par
		\State \hskip\algorithmicindent 
		$\psi^{(i+1)} \gets \psi^{(i)} - 
		\lambda_\psi \nabla_\psi \mathcal{L}_{mm}(\psi^{(i)})$

		\State Sample $\mathcal D$ and train action policy:
		\State \hskip\algorithmicindent
			evaluate $R_\psi^{(i+1)} = \mathcal{L}_{mm}(\psi^{(i)}) - 
			\mathcal{L}_{mm}(\psi^{(i+1)})$  \par 
		\State \hskip\algorithmicindent 
			evaluate $\hat{V}^\prime = \hat{V}(\bs{s}^\prime; \nu^{(j+M)})$ \par
		\State \hskip\algorithmicindent 
			$w \gets \pi(\bs{a}|\bs{s};\varphi^{(k)}) / {\pi(\bs{a}|\bs{s};\varphi^{(<k)})}$ 
		\State \hskip\algorithmicindent 
			$\varphi^{(k+1)} \gets \varphi^{(k)} + \lambda^\varphi \nabla_\varphi w \mathcal{L}_{ap}(\varphi^{(k)})$
			given $R_\psi^{(i+1)}$, $\hat{V}^\prime$
		\EndFor
	\end{algorithmic}
\end{algorithm}

\begin{algorithm}[H]
	\caption{Fitted Policy Evaluation (cf. \citet{svg})}
	\label{alg:fpeval}
	\begin{algorithmic}[1]
		\State \textbf{Given} \par 
		\hskip\algorithmicindent outer loop time $t$ \par 
		\hskip\algorithmicindent experience pool $\mathcal D$ \par 
		\hskip\algorithmicindent value function $V(\bs{s}; \nu^{(j)})$ \par 
		\hskip\algorithmicindent gradient step counter $i$, $j$, $k$
		\State Clone parameter $\tilde\nu \gets \nu^{(j)}$
		\For {$m = 1\dots M$}
		\State Sample 
			$\left(\bs{s}^\tau, \bs{a}^\tau, \pi(\bs{a}^\tau|\bs{s}^\tau; \varphi^{(<k)}), 	\bs{s}^{\tau+1}\right)$ 
			from $\mathcal D$ ($\tau < t$)
		\State Evaluate 
			$R_\psi^{(i+1)}=\mathcal{L}_{mm}(\psi^{(i)}) - \mathcal{L}_{mm}(\psi^{(i+1)})$
		\State $y = R_\psi^{(i+1)} + \gamma \hat{V}(\bs{s}^{\tau+1}; \tilde\nu)$ 
		\State $w = \pi(\bs{a}^\tau|\bs{s}^\tau;\varphi^{(k)}) / {\pi(\bs{a}^\tau|\bs{s}^\tau;\varphi^{(<k)})}$
		\State Apply updates 
			$\nu^{(j+m)} \gets \nu^{(j+m-1)} - 
			\nabla_\nu \frac w2 \left( y - V(\bs{s}; \nu^{(j+m-1)})\right)^2$
		\State Every $C$ updates, $\tilde\nu \gets \nu^{(j+m)}$
		\EndFor
	\end{algorithmic}
\end{algorithm}

\end{document}